\documentclass{article}
\usepackage{ijcai19}

\usepackage{times}
\usepackage{soul}
\usepackage{url}
\usepackage[hidelinks]{hyperref}
\usepackage[utf8]{inputenc}
\usepackage[small]{caption}
\usepackage{graphicx}
\usepackage{amsmath}
\usepackage{booktabs}
\usepackage{algorithm}
\usepackage{algorithmic}

\usepackage{graphicx}
\usepackage{latexsym}
\usepackage{amsmath}
\usepackage{amssymb}
\usepackage{subfigure}
\usepackage{xcolor}
\usepackage{enumerate}
\usepackage{comment}
\usepackage{subfigure}
\usepackage{multirow}
\usepackage{bm}
\usepackage{wrapfig}

\title{3{D}ViewGraph: Learning Global Features for 3D Shapes from A Graph of Unordered Views with Attention}

\author{
Zhizhong Han\textsuperscript{1,2}, Xiyang Wang\textsuperscript{1}, Chi-Man Vong\textsuperscript{3}, Yu-Shen Liu\textsuperscript{1}\thanks{Corresponding author: Yu-Shen Liu}, Matthias Zwicker\textsuperscript{2},C.L. Philip Chen\textsuperscript{4}\\
\textsuperscript{1}School of Software, Tsinghua University, Beijing, China \\
Beijing National Research Center for Information Science and Technology (BNRist)\\
\textsuperscript{2}Department of Computer Science, University of Maryland, College Park, USA\\
\textsuperscript{3}Department of Computer and Information Science, University of Macau, Macau, China\\
\textsuperscript{4}Faculty of Science and Technology, University of Macau, Macau, China\\
h312h@umd.edu,
wangxiya16@mails.tsinghua.edu.cn,
cmvong@um.edu.mo
liuyushen@tsinghua.edu.cn,
zwicker@cs.umd.edu,
philip.chen@ieee.org
}

\begin{document}
\maketitle

\begin{abstract}
  Learning global features by aggregating information over multiple views has been shown to be effective for 3D shape analysis. For view aggregation in deep learning models, pooling has been applied extensively. However, pooling leads to a loss of the content within views, and the spatial relationship among views, which limits the discriminability of learned features. We propose \textit{3DViewGraph} to resolve this issue, which learns 3D global features by more effectively aggregating unordered views with attention. Specifically, unordered views taken around a shape are regarded as view nodes on a view graph. 3DViewGraph first learns a novel latent semantic mapping to project low-level view features into meaningful latent semantic embeddings in a lower dimensional space, which is spanned by latent semantic patterns. Then, the content and spatial information of each pair of view nodes are encoded by a novel spatial pattern correlation, where the correlation is computed among latent semantic patterns. Finally, all spatial pattern correlations are integrated with attention weights learned by a novel attention mechanism. This further increases the discriminability of learned features by highlighting the unordered view nodes with distinctive characteristics and depressing the ones with appearance ambiguity. We show that 3DViewGraph outperforms state-of-the-art methods under three large-scale benchmarks.
\end{abstract}

\section{Introduction}
% The very first letter is a 2 line initial drop letter followed
% by the rest of the first word in caps.
%
% form to use if the first word consists of a single letter:
% \IEEEPARstart{A}{demo} file is ....
%
% form to use if you need the single drop letter followed by
% normal text (unknown if ever used by the IEEE):
% \IEEEPARstart{A}{}demo file is ....
%
% Some journals put the first two words in caps:
% \IEEEPARstart{T}{his demo} file is ....
%
% Here we have the typical use of a "T" for an initial drop letter
% and "HIS" in caps to complete the first word.
Global features of 3D shapes can be learned from raw 3D representations, such as meshes, voxels, and point clouds.
%MZ: commented this out since it is repetitive with the following sentence
%, and 2D views~\cite{su15mvcnn, Bshi2015, su16mvcnn, MvCNN2017, Takahiko16, chuwang2017}.
As an alternative, a number of works in 3D shape analysis employed multiple views~\cite{su15mvcnnijcai,Zhizhong2019seq} as raw 3D representation, exploiting the advantage that multiple views can facilitate understanding of both manifold and non-manifold 3D shapes via computer vision techniques. Therefore, effectively and efficiently aggregating comprehensive information over multiple views, is critical for the discriminability of learned features, especially in deep learning models.

Pooling was designed as a procedure for information abstraction in deep learning models. In order to describe a 3D shape by considering features from multiple views, view aggregation is usually performed by max or mean pooling, where pooling only employs the max or mean value of each dimension across all view features~\cite{su15mvcnnijcai}. Although pooling is able to eliminate the rotation effect of 3D shapes, both the content information within views and the spatial relationship among views cannot be fully preserved. As a consequence, this limits the discriminability of learned features. In this work, we address the challenge to learn 3D features in a deep learning model by more effectively aggregating the content information within individual views, and the spatial relationship among multiple unordered views.

To tackle this issue, we propose a novel deep learning model called \textit{3D View Graph} (3DViewGraph), which learns 3D global features from multiple unordered views. By taking multiple views around a 3D shape on a unit sphere, we represent the shape as a view graph formed by the views, where each view denotes a node, and the nodes are fully connected with each other by edges. 3DViewGraph learns highly discriminative global 3D shape features
%from its corresponding view graph
by simultaneously encoding both the content information within the view nodes, and the spatial relationship among the view nodes.

\begin{enumerate}[i)]
\item We propose a novel deep learning model called 3DViewGraph for 3D global feature learning by effectively aggregating multiple unordered views. It not only encodes the content information within all views, but also preserves the spatial relationship among the views.
\item We propose an approach to learn a low-dimensional latent semantic embedding of the views by directly capturing the similarities between each view and a set of latent semantic patterns. As an advantage, 3DViewGraph avoids mining the latent semantic patterns across the whole training set explicitly.

%We develop a lower-dimensional, latent semantic view representation, which directly captures the similarities between views and a set of latent semantic patterns. As an advantage, our approach avoids mining the latent semantic patterns across the whole training set explicitly.

\item We perform view aggregation by integrating a novel spatial pattern correlation, which encodes the content information and the spatial relationship in each pair of views.
\item We propose a novel attention mechanism to increase the discriminability of learned features by highlighting the unordered view nodes with distinctive characteristics and depressing the ones with appearance ambiguities.
\end{enumerate}

\section{Related work}
\label{sec:relatedworks}
Deep learning models have made a big progress on learning 3D shape features from different raw representations, such as meshes~\cite{HanTIP18ijcai}, voxels~\cite{3dganWuijcai}, point clouds~\cite{nipspoint17ijcai} and views~\cite{su15mvcnnijcai}. Because of page limit, we focus on reviewing view-based deep learning models to highlight the novelty of our view aggregation.

\noindent \textbf{View-based methods. }View-based methods represent a 3D shape as a set of rendered views~\cite{AsakoCVPR2018} or panorama views~\cite{Sfikas17ijcai}. Besides direct set-to-set comparison~\cite{tmmbs2016ijcai}, pooling is the widely used way of aggregating multiple views in deep learning models~\cite{su15mvcnnijcai}. In addition to global feature learning, pooling can also be used to learn local features~\cite{MvCNN2017,Yu_2018_CVPR} for segmentation or correspondence by aggregating local patches.

Although pooling can aggregate views on the fly in the models, it can not encode all the content information within views and the spatial relationship among views. Thus, the strategies of concatenation~\cite{Savva2016SHRECijcai}, view pair weighting \cite{JohnsLD16}, cluster specified pooling~\cite{chuwang2017ijcai}, RNN~\cite{Zhizhong2018seqijcai}, were employed to resolve this issue. However, these methods can not learn from unordered views or fully capture the spatial information among unordered views.

To resolve the aforementioned issues, 3DViewGraph aggregates unordered views more effectively by simultaneously encoding their content information and spatial relationship.

\noindent \textbf{Graph-based methods. }To handle the irregular structure of graphs, various methods have been proposed~\cite{HamiltonYL17ijcai}. Although we formulate the multiple views from a 3D shape as a view graph, existing methods proposed for graphs cannot be directly used for learning the 3D feature in our scenario. The reasons are two-fold. First, these methods mainly focus on how to locally learn meaningful representation for each node in a graph from its raw attributes rather than globally learning the feature of the whole graph. Second, these methods mainly learns how to process the nodes in a graph with firm order, while the order of views involved in 3DViewGraph are always ambiguous because of the rotation of 3D shapes.

Moreover, some methods have employed graphs to retrieve 3D shapes from multiple views~\cite{Liu2015GCV,AnnLiu16}. Different from these methods, 3DViewGraph employs a more efficient way of view aggregation in deep learning models, which makes the learned features useful for both classification and retrieval.

\section{3DViewGraph}
\label{sec:model}

\begin{figure*}[!htb]
  \centering
  % the following command controls the width of the embedded PS file
  % (relative to the width of the current column)
  %\includegraphics[width=.95\linewidth, bb=39 696 126 756]{figures/definition3.eps}
   \includegraphics[width=\linewidth]{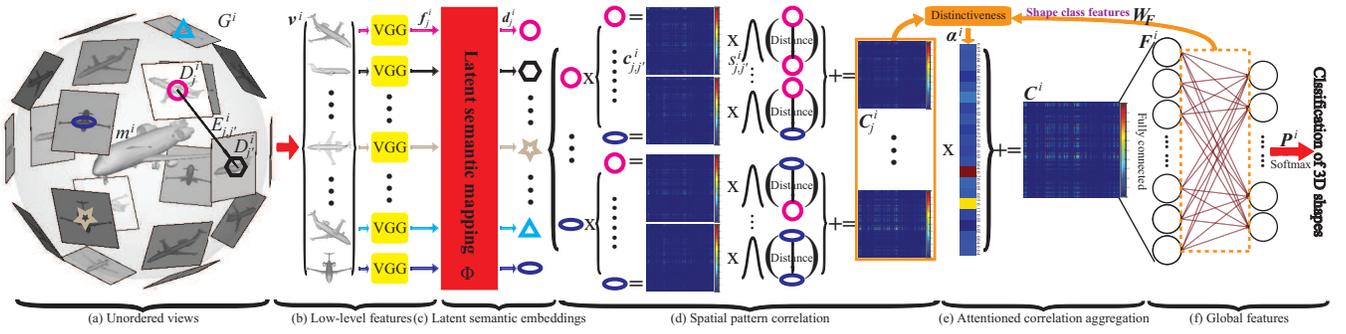}
  % replacing the above command with the one below will explicitly set
  % the bounding box of the PS figure to the rectangle (xl,yl),(xh,yh).
  % It will also prevent LaTeX from reading the PS file to determine
  % the bounding box (i.e., it will speed up the compilation process)
  % \includegraphics[width=.95\linewidth, bb=39 696 126 756]{sampleFig}
  %
  %
\caption{\label{fig:Framework} The demonstration of 3DViewGraph framework. }
\end{figure*}

\noindent\textbf{Overview. }Fig.~\ref{fig:Framework} shows an overview of 3DViewGraph, where the global feature $\bm{F}^i\in\mathbb{R}^{1\times F}$ of a 3D shape $m^i$ is learned from its corresponding view graph $G^i$. Here, $m^i$ is the $i$-th shape in a training set of $M$ 3D shapes, where $i\in[1,M]$. Based on the $F$-dimensional feature $\bm{F}^i$, 3DViewGraph classifies $m^i$ into one of $L$ shape classes according to the probability $\bm{P}^i=[P^i(l^i=1|\bm{F}^i),...,P^i(l^i=a|\bm{F}^i),...,P^i(l^i=L|\bm{F}^i)]$, which is provided by a final softmax classifier (Fig.~\ref{fig:Framework}(f)), where $l^i$ is the class label of $m^i$.

We first take a set of unordered views $\bm{v}^i=\{v_j^i|j\in[1,V]\}$ on a unit sphere centered at $m^i$, as shown in Fig.~\ref{fig:Framework}(a). Here, we use ``unordered views'' to  indicate that the views cannot be organized in a sequential way. The views $v_j^i$ are regarded as view nodes $D_j^i$ (briefly shown by symbols) of an undirected graph $G^i$, where each $D_j^i$ is fully connected with other view nodes $D_{j'}^i$ by edges $E_{j,j'}^i$, such that $G^i=(\{D_j^i\},\{E_{j,j'}^i\})$.

Next, we extract low-level features $\bm{f}_j^i$ of each view $v_j^i$ using a fine-tuned VGG19 network~\cite{Simonyan14c}, as shown in Fig.~\ref{fig:Framework}(b), where $\bm{f}_j^i\in\mathbb{R}^{1\times 4096}$ is extracted from the last fully connected layer. To obtain lower-dimensional, semantically more meaningful view features, we subsequently learn a latent semantic mapping $\Phi$ (Fig.~\ref{fig:Framework}(c)) to project a low-level view feature $\bm{f}_j^i$ into
% latent semantic space as
its latent semantic embedding $\bm{d}_j^i$.

To resolve the effect of rotation, 3DViewGraph encodes the content and spatial information of $G^i$ by exhaustively computing our novel spatial pattern correlation between each pair of view nodes. As illustrated in Fig.~\ref{fig:Framework}(d), we compute the pattern correlation $\bm{c}_{j,j'}^i$ between $D_j^i$ and each other node $D_{j'}^i$, and we weight it with their spatial similarity $s_{j,j'}^i$. %, which forms spatial pattern correlation.
In addition, for each node $D_j^i$, we compute its cumulative correlation $\bm{C}_{j}^i$ to summarize all spatial pattern correlations as the characteristics of the 3D shape from the $j$-th view node $D_j^i$.
%starting from $D_j^i$
%to represent the characteristics of 3D shape $m^i$ from the $j$-th view node $D_j^i$ on $G^i$.

Finally, we obtain the global feature $\bm{F}^i$ of shape $m^i$ by integrating all cumulative correlations $\bm{C}_{j}^i$ with our novel attention weights $\bm{\alpha}^i$, as shown in Fig.~\ref{fig:Framework}(e) and (f). $\bm{\alpha}^i$ aims to highlight the view nodes with distinctive characteristics while depressing the ones with appearance ambiguity.

%\subsection{Unordered view capturing}
%We take $V$ unordered views $v_j^i$ around the $i$-th 3D shape $m^i$ on a unit sphere, and the view set $\bm{v}^i$ formed by these views are shown in Fig.~\ref{fig:Framework}(a), where $j\in[1,V]$. The views $v_j^i$ in $\bm{v}^i$ are uniformly distributed on the sphere in an unordered manner, fully covering the 3D shape $m^i$ including the top and bottom of the sphere. In this work, we place the cameras at the 20 vertices of a regular dodecahedron, which are located on the unit sphere, such that $V=20$. We render each 3D shape with OpenGL using perspective projection, a diffuse shading model, and a single directional light source.

\noindent\textbf{Latent semantic mapping learning. }To learn global features from unordered views, 3DViewGraph encodes the content information within all views and the spatial relationship among views in a pairwise way. 3DViewGraph relies on the intuition that correlations between pairs of views can effectively represent discriminative characteristics of a 3D shape, especially considering the relative spatial position of the views. To implement this intuition, each view should be encoded in terms of a small set of common elements across all views in  the training set. Unfortunately, the low-level features $\bm{f}_j^i$ are too high dimensional and not suitable as a representation of the views in terms of a set of common elements.
%Unfortunately, although the low-level feature $\bm{f}_j^i$ could represent each view $v_j^i$ by messy descriptions, $\bm{f}_j^i$ is with high dimension and not capable of representing $v_j^i$ by common elements across all views in the training set.

\begin{figure}[htb]
  \centering
  % the following command controls the width of the embedded PS file
  % (relative to the width of the current column)
  %\includegraphics[width=.95\linewidth, bb=39 696 126 756]{figures/definition3.eps}
   \includegraphics[width=\linewidth]{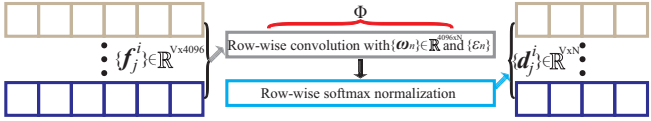}
  % replacing the above command with the one below will explicitly set
  % the bounding box of the PS figure to the rectangle (xl,yl),(xh,yh).
  % It will also prevent LaTeX from reading the PS file to determine
  % the bounding box (i.e., it will speed up the compilation process)
  % \includegraphics[width=.95\linewidth, bb=39 696 126 756]{sampleFig}
  %
  %
\caption{\label{fig:mapping} The demonstration of latent semantic mapping $\Phi$.}
\end{figure}

To resolve this issue, 3DViewGraph introduces a latent semantic mapping $\Phi$ by learning a kernel function $K$ to directly capture the similarities between $V$ views $v_j^i$ and $N$ latent semantic patterns $\{\bm{\phi}_n\}$. Our approach avoids additionally and explicitly mining $\{\bm{\phi}_n\}$ across the whole training set as the common elements. $\Phi$ projects low-level view features $\bm{f}_j^i$ into latent semantic space spanned by $\{\bm{\phi}_n\}$ as latent semantic emdeddings $\bm{d}_j^i$. $\bm{d}_j^i$ represents view nodes $D^i_j$ with more semantic meaning but lower dimension than $\bm{f}_j^i$. Specifically, predicted by kernel $K$, the $n$-th dimension of $\bm{d}_j^i$ characterizes the similarity between $\bm{f}_j^i$ and the $n$-th semantic pattern $\bm{\phi}_n$, such that $\bm{d}_j^i=[K(\bm{f}^i_j,\bm{\phi}_1),...,K(\bm{f}^i_j,\bm{\phi}_n),...,K(\bm{f}^i_j,\bm{\phi}_N)]\in \mathbb{R}^{1\times N}$. We define the kernel $K$ as

%\begin{equation}
%\label{eq:semanticembedding}
%\bm{d}_j^i(n)=\frac{e^{-\beta\|\bm{f}_j^i-\bm{\phi}_n\|_2^2}}{\sum_{n'=1}^{N}e^{-\beta\|\bm{f}_j^i-\bm{\phi}_{n'}\|_2^2}},
%\end{equation}

\begin{equation}
\label{eq:semanticembedding}
K(\bm{f}_j^i,\bm{\phi}_n)=\frac{exp(-\beta\|\bm{f}_j^i-\bm{\phi}_n\|_2^2)}{\sum_{n'=1}^{N}exp(-\beta\|\bm{f}_j^i-\bm{\phi}_{n'}\|_2^2)},
\end{equation}

\noindent where the similarity $K(\bm{f}_j^i,\bm{\phi}_n)$ is inversely proportional to the distance between $\bm{f}_j^i$ and $\bm{\phi}_n$ through $exp()$, and gets normalized across the similarities between $\bm{f}_j^i$ and all $\bm{\phi}_n$. Parameter $\beta$ controls the decay of the response with the distance. This equation can be further simplified by cancelling the norm of $\bm{f}_j^i$ from the numerator and the denominator as follows,

%\begin{equation}
%\label{eq:semanticembeddings}
%\begin{split}
%\bm{d}_j^i(n)=\frac{e^{-\beta\|\bm{f}_j^i\|_2^2+2\beta\bm{f}_j^i\bm{\phi}_n^{\mathrm{T}}-\beta\|\bm{\phi}_n\|_2^2}}{\sum_{n'=1}^{N}e^{-\beta\|\bm{f}_j^i\|_2^2+2\beta\bm{f}_j^i\bm{\phi}_{n'}^{\mathrm{T}}-\beta\|\bm{\phi}_{n'}\|_2^2}} \\
%=\frac{e^{2\beta\bm{f}_j^i\bm{\phi}_n^{\mathrm{T}}-\beta\|\bm{\phi}_n\|_2^2}}{\sum_{n'=1}^{N}e^{2\beta\bm{f}_j^i\bm{\phi}_{n'}^{\mathrm{T}}-\beta\|\bm{\phi}_{n'}\|_2^2}} \\
%=\frac{e^{\bm{f}_j^i\bm{\omega}_n+\varepsilon_n}}{\sum_{n'=1}^{N}e^{\bm{f}_j^i\bm{\omega}_{n'}+\varepsilon_{n'}}},
%\end{split}
%\end{equation}

\begin{equation}
\label{eq:semanticembeddings}
\begin{split}
& K(\bm{f}_j^i,\bm{\phi}_n)=\frac{exp(-\beta\|\bm{f}_j^i\|_2^2+2\beta\bm{f}_j^i\bm{\phi}_n^{\mathrm{T}}-\beta\|\bm{\phi}_n\|_2^2)}{\sum_{n'=1}^{N}exp(-\beta\|\bm{f}_j^i\|_2^2+2\beta\bm{f}_j^i\bm{\phi}_{n'}^{\mathrm{T}}-\beta\|\bm{\phi}_{n'}\|_2^2)} \\
& \textcolor{white}{K(\bm{f}_j^i,\bm{\phi}_n)}=\frac{exp(2\beta\bm{f}_j^i\bm{\phi}_n^{\mathrm{T}}-\beta\|\bm{\phi}_n\|_2^2)}{\sum_{n'=1}^{N}exp(2\beta\bm{f}_j^i\bm{\phi}_{n'}^{\mathrm{T}}-\beta\|\bm{\phi}_{n'}\|_2^2)} \\
& \textcolor{white}{K(\bm{f}_j^i,\bm{\phi}_n)}=\frac{exp(\bm{f}_j^i\bm{\omega}_n+\varepsilon_n)}{\sum_{n'=1}^{N}exp(\bm{f}_j^i\bm{\omega}_{n'}+\varepsilon_{n'})},
\end{split}
\end{equation}

\noindent where in the last step, we substituted $2\beta\bm{\phi}_n^{\mathrm{T}}$ and $-\beta\|\bm{\phi}_n\|_2^2$ by $\bm{\omega}_n$ and $\varepsilon_n$, respectively. Here, $\{\bm{\omega}_n\}$, $\{\varepsilon_n\}$ and $\{\bm{\phi}_n\}$ are sets of learnable parameters, in addition, both $\{\bm{\omega}_n\}$ and $\{\varepsilon_n\}$ depend on $\{\bm{\phi}_n\}$. However, to obtain more flexible training by following the viewpoint in~\cite{ArandjelovicGTP16ijcai}, we employ two independent sets of $\{\bm{\omega}_n\}$, $\{\varepsilon_n\}$,
%and $\{\bm{\phi}_n\}$
decoupling $\{\bm{\omega}_n\}$ and $\{\varepsilon_n\}$ from $\{\bm{\phi}_n\}$. This decoupling enables 3DViewGraph to directly predict the similarity between $\bm{f}_j^i$ and $\bm{\phi}_n$ by the kernel $K$ without explicitly mining $\bm{\phi}_n$ across all low-level view features in the training set.

Based on the last line in Eq.~\ref{eq:semanticembeddings}, we implement the latent semantic mapping $\Phi$ as a row-wise convolution with each pair of $\{\bm{\omega}_n\}$ and $\{\varepsilon_n\}$ corresponding to a filter and a row-wise softmax normalization, as shown in Fig.~\ref{fig:mapping}.

%The  intuition  behindrepresenting  view  nodes  in  terms  of  latent  semantic  patternsis that this will facilitate capturing correlations between pairsof view nodes across these patterns, and we will introduce ourapproach to compute these correlations next.

%It is meaningful to implicitly employ the semantic patterns to represent view nodes in the latent semantic space, since this would facilitate 3DViewGraph to capture the correlation between each pair of view nodes across all semantic patterns. This will be introduced in detail in the subsequent subsection.

%In addition, a single layer perceptron can be used to learn the semantic patterns $\{\bm{\phi}_n\}$ which form the weights of the single layer perceptron.

%\subsection{Spatial pattern correlation}

%Spatial pattern correlation is conducted on each pair of view nodes $D_{j}^i$ and $D_{j'}^i$ to represent some characteristics of 3D shape $m^i$. In this procedure, the pattern correlation $\bm{c}_{j,j'}^i$ between the corresponding latent semantic embeddings $\bm{d}_j^i$ and $\bm{d}_{j'}^i$ is weighted by their spatial similarity $s_{j,j'}^i$.

\noindent\textbf{Spatial pattern correlation. }The pattern correlation $\bm{c}_{j,j'}^i$ aims to encode the content of view nodes $D_{j}^i$ and $D_{j'}^i$. $\bm{c}_{j,j'}^i$ makes the semantic patterns that co-occur in both views more prominent while the non-co-occurring ones more subtle. More precisely, we use the latent semantic embeddings $\bm{d}_j^i$ and $\bm{d}_{j'}^i$ to compute $\bm{c}_{j,j'}^i$ as follows,

%The pattern correlation $\bm{c}_{j,j'}^i$ represents the correlation across all semantic patterns between a pair of view nodes $D_{j}^i$ and $D_{j'}^i$. Intuitively, $\bm{c}_{j,j'}^i$ makes the semantic patterns that co-occur in both views more prominent, while the non-co-occurring ones are subtle. In other words, the correlations emphasize the characteristics of 3D shape $m^i$ from different perspectives. More precisely, we use the latent semantic embeddings $\bm{d}_j^i$ and $\bm{d}_{j'}^i$ to compute $\bm{c}_{j,j'}^i$ as follows,

\begin{equation}
\label{eq:semanticcorrelation}
\bm{c}_{j,j'}^i=(\bm{d}_j^i)^{\mathrm{T}}\times\bm{d}_{j'}^i,
\end{equation}

\noindent where $\bm{c}_{j,j'}^i$ is a $N\times N$ dimensional matrix whose entry $\bm{c}_{j,j'}^i(n,n')$ measures the correlation between the semantic pattern $\bm{\phi}_n$ contributing to $\bm{d}_j^i$ and $\bm{\phi}_{n'}$ contributing to $\bm{d}_{j'}^i$.

\begin{figure}[htb]
  \centering
  % the following command controls the width of the embedded PS file
  % (relative to the width of the current column)
  %\includegraphics[width=.95\linewidth, bb=39 696 126 756]{figures/definition3.eps}
   \includegraphics[width=\linewidth]{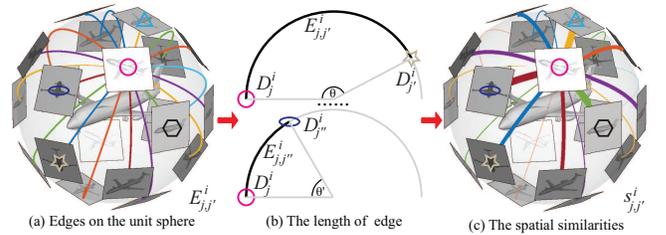}
  % replacing the above command with the one below will explicitly set
  % the bounding box of the PS figure to the rectangle (xl,yl),(xh,yh).
  % It will also prevent LaTeX from reading the PS file to determine
  % the bounding box (i.e., it will speed up the compilation process)
  % \includegraphics[width=.95\linewidth, bb=39 696 126 756]{sampleFig}
  %
  %
\caption{\label{fig:spatial} The illustration of spatial similarity $s^i_{j,j'}$.}
\end{figure}

We further enhance the pattern correlation $\bm{c}_{j,j'}^i$ between the view nodes $D_{j}^i$ and $D_{j'}^i$ by their spatial similarity $s_{j,j'}^i$, which forms the spatial pattern correlation $s_{j,j'}^i\bm{c}_{j,j'}^i$.
%between $D_{j}^i$ and $D_{j'}^i$.

Fig.~\ref{fig:spatial} visualizes how we compute the spatial similarity $s_{j,j'}^i$. In Fig.~\ref{fig:spatial}(a), we show all edges $E_{j,j'}^i$ connecting $D_j^i$ to all other view nodes $D_{j'}^i$ in different colors, where $D_j^i$ is briefly shown by symbols. The length of $E_{j,j'}^i$ is measured by the length of the shortest arc connecting the two view nodes $D_{j}^i$ and $D_{j'}^i$ on the unit sphere. Thus, $E_{j,j'}^i=2\pi\times 1 \times(\theta/2\pi)=\theta$ as illustrated in Fig.~\ref{fig:spatial}(b), where $\theta$ is the central angle of the arc and the factor $1$ corresponds to the radius of the unit sphere. %However, to avoid the discriminability of learned features from being affected by
To reduce the high variance of $\{E_{j,j'}^i\}$, we employ $E_{j,j'}^i=0.5(1-\cos\theta)$ instead of $E_{j,j'}^i=\theta$,
%. $0.5(1-\cos\theta)$
which normalizes $E_{j,j'}^i$ into the range of $[0,1]$. Finally, $s_{j,j'}^i$ is inversely proportional to $E_{j,j'}^i$ as follows,

% with more appropriate variance than normalized $\theta$.

%In addition, we do not employ the nearest geodesic distance on $G^i$ which measures $E_{j,j'}^i$ by all edges in the nearest path from $D_j^i$ to $D_{j'}^i$, since the nearest geodesic distance on $G^i$ would not be accuracy because of the small number of view nodes (to prevent 3DViewGraph from learning from redundant information).

%We model the spatial similarity $s_{j,j'}^i$ between view nodes $D_j^i$ and $D_{j'}^i$ by a Gauss function with $E_{j,j'}^i$ as input, which makes $s_{j,j'}^i$ inversely proportional to $E_{j,j'}^i$ as defined below,

\begin{equation}
\label{eq:spatialsimilarity}
s_{j,j'}^i=exp(-\sigma E_{j,j'}^i),
\end{equation}

\noindent where $\sigma$ is a parameter to control the decay of the response with the edge length. In Fig.~\ref{fig:spatial}(c), we visualize $s_{j,j'}^i$ by mapping the value of $s_{j,j'}^i$ to the width of edges $E^i_{j,j'}$.%, where the edge colors correspond to the ones in \ref{fig:spatial} (a).

To represent the characteristics of 3D shape $m^i$ from the $j$-th view node $D_j^i$ on $G^i$, we finally introduce the cumulative correlation $\bm{C}_{j}^i$, which encodes all spatial pattern correlations starting from $D_j^i$ as follows,

\begin{equation}
\label{eq:summarizedsemanticcorrelation}
\bm{C}_{j}^i=\sum_{j'=1}^V s_{j,j'}^i\bm{c}_{j,j'}^i.
\end{equation}

\begin{figure}[htb]
  \centering
  % the following command controls the width of the embedded PS file
  % (relative to the width of the current column)
  %\includegraphics[width=.95\linewidth, bb=39 696 126 756]{figures/definition3.eps}
   \includegraphics[width=\linewidth]{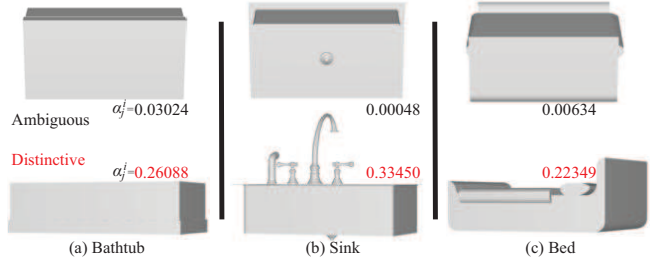}
  % replacing the above command with the one below will explicitly set
  % the bounding box of the PS figure to the rectangle (xl,yl),(xh,yh).
  % It will also prevent LaTeX from reading the PS file to determine
  % the bounding box (i.e., it will speed up the compilation process)
  % \includegraphics[width=.95\linewidth, bb=39 696 126 756]{sampleFig}
  %
  %
\caption{\label{fig:ambiguity} The ambiguous views and distinctive views.}
\end{figure}

\noindent\textbf{Attentioned correlation aggregation. }Intuitively, more views will provide more information to any deep learning model, which should allow it to produce more discriminative 3D features. However, additional views may also introduce appearance ambiguities that negatively affect the discriminability of learned features, as shown in Fig.~\ref{fig:ambiguity}.

To resolve this issue, 3DViewGraph employs a novel attention mechanism in the aggregation of the 3D shape characteristics from all unordered view nodes of a shape, as illustrated in Fig.~\ref{fig:Framework}(e). 3DViewGraph learns attention weights $\bm{\alpha}^i=\{\alpha_j^i|j\in[1,V]\}$ for all view nodes $D_j^i$ on $G^i$, where $\alpha_j^i$ would be a large value (the second row in Fig.~\ref{fig:ambiguity}) if the view $v_j^i$ has distinctive characteristics, while $\alpha_j^i$ would be a small value (the first row in Fig.~\ref{fig:ambiguity}) if $v_j^i$ exhibits appearance ambiguity with views from other shapes. Note that $\sum_{j=1}^V\alpha_j^i=1$.

Our novel attention mechanism evaluates how distinctive each view is to the views that 3DViewGraph has processed. To comprehensively represent the characteristics of the views that 3DViewGraph has processed, the attention mechanism employs the fully connected weights $\bm{W}_F$ in the final softmax classifier which accumulates the information of all views, as shown in Fig.~\ref{fig:Framework}(f). The attention mechanism projects the characteristics $\bm{C}_{j}^i$ of 3D shape $m^i$ from the $j$-th view node $D^i_j$ and the characteristics $\bm{W}_F$ of the views that 3DViewGraph has processed into a common space to calculate the distinctiveness of view $v_j^i$, as defined below,

\begin{equation}
\label{eq:attention}
\begin{aligned}
& \alpha_j^i=\bm{\omega}(\bm{W}_C\bm{C}_j^i\bm{\omega}_C+\bm{W}_F\bm{\omega}_F+\bm{b}),\\
& \bm{\alpha}^i=softmax(\bm{\alpha}^i),\\
\end{aligned}
\end{equation}

\noindent where $\bm{W}_C$, $\bm{\omega}_C$, $\bm{\omega}_F$, $\bm{b}$ and $\bm{\omega}$ are learnable parameters in the attention mechanism, $\bm{W}_F\in\mathbb{R}^{L\times F}$, where $F$ is the dimension of the learned global feature $\bm{F}^i$, and $L$ is the number of shape classes. With $\bm{W}_C\in\mathbb{R}^{L\times V}$ and $\bm{\omega}_C\in\mathbb{R}^{V\times 1}$, $\bm{C}_j^i$ is projected into a $L\times 1$ dimensional space, where $\bm{b}\in\mathbb{R}^{L\times 1}$ is a bias in that space. In addition, $\bm{W}_F$ is projected into the same space by $\bm{\omega}_F\in\mathbb{R}^{K\times 1}$ to compute the similarities between $\bm{C}_j^i$ and $\bm{W}_F$ along all $L$ dimensions. Subsequently, the attention weight $\alpha_j^i$ is calculated by comprehensively summarizing all similarities along all the $L$ dimensions with a linear mapping $\bm{\omega}\in\mathbb{R}^{1\times L}$. Finally, the $\alpha_j^i$ in $\bm{\alpha}^i$ for all views of $i$-th shape are normalized by softmax normalization.

Based on $\bm{\alpha}^i$, the characteristics $\bm{C}_{j}^i$ of 3D shape $m^i$ from all view nodes are aggregated with weighting $\bm{\alpha}^i$ into attentioned correlation aggregation $\bm{C}^i$, as defined below,

\begin{equation}
\label{eq:attentionaggregation1}
\bm{C}^i=\sum_{j=1}^V \alpha_j^i\bm{C}_{j}^i,
\end{equation}

\noindent where $\bm{C}^i$ represents 3D shape $m^i$ as a $N\times N$ matrix, as shown in Fig.~\ref{fig:Framework}(e). Finally, the global feature $\bm{F}^i$ of 3D shape $m^i$ is learned by a fully connected layer with attentioned correlation aggregation $\bm{C}^i$ as input, as shown in Fig.~\ref{fig:Framework}(f), where the fully connected layer is followed by a sigmoid function.

%Here, $\bm{F}^i\in\mathbb{R}^{1\times F}$ is a $F$ dimensional vector.

Using $\bm{F}^i$, the final softmax classifier computes the probabilities $\bm{P}^i$ to classify the 3D shape $m^i$ into one of $L$ shape classes as %defined as follows,

\begin{equation}
\label{eq:attentionaggregation}
\bm{P}^i=softmax(\bm{W}_F \bm{F}^i+\bm{b}_F),
\end{equation}

\noindent where $\bm{W}_F\in \mathbb{R}^{L\times K}$ and $\bm{b}_F\in\mathbb{R}^{K\times 1}$ are learnable parameters for the computation of $\bm{P}^i$. $\bm{W}_F$ is used to represent all the characteristics of views that 3DViewGraph has processed, as employed to calculate $\bm{\alpha}^i$ in Eq.~\ref{eq:attention}.

\noindent\textbf{Learning inference. }The parameters involved in 3DViewGraph are optimized by minimizing the log-likelihood $O$ over $M$ 3D shapes in the training set, where $\bm{Q}^i$ is the truth label, %as defined below,

%\begin{equation}
%\label{eq:viewatten7}
%O=-\frac{1}{M}\sum_{i=1}^M\sum_{a=1}^L \log P^i(l^i=a).
%\end{equation}

%\begin{equation}
%\label{eq:viewatten7}
%O=-\frac{1}{M}\sum_{i=1}^M \bm{Q}^i\log \bm{P}^i.
%\end{equation}

\begin{equation}
\label{eq:viewatten7}
O=-\frac{1}{M}\sum_{i=1}^M\sum_{a=1}^L Q^i(l^i=a)\log P^i(l^i=a).
\end{equation}

%\noindent \textbf{Learning inference.}
%\noindent\textbf{Learning inference.}
The parameter optimization is conducted by back propagation of classification errors of 3D shapes. Noteworthy, $\bm{W}_F$ is updated by two elements with the learning rate $\varepsilon$ as follows,

\begin{equation}
\label{eq:sga}
\bm{W}_F\gets \bm{W}_F-\varepsilon(\frac{\partial O}{\partial \bm{W}_F}+\sum_{j=1}^V\frac{\partial \alpha_j^i}{\partial \bm{W}_F}).
\end{equation}

The advantage of Eq.~(\ref{eq:sga}) is that $\bm{W}_F$ can be learned more flexibly for optimization convergence. $\bm{W}_F$ also enables $\bm{\alpha}^i$ to simultaneously observe the characteristics of shape $m^i$ from each view node $D_j^i$ and take all views that have been processed from different shapes as reference.

\section{Results and analysis}
\label{sec:results}
We evaluate 3DViewGraph by comparing it with the state-of-the-art methods in shape classification and retrieval under ModelNet40~\cite{Wu2015ijcai}, ModelNet10 and ShapeNetCore55~\cite{3dor20171050ijcai}. We also show ablation studies to justify the effectiveness of novel elements.

\begin{table}[h]
\centering
%\caption{The number of row-wise kernels and droupout ratio comparison under ModelNet40, $\varepsilon$=2e-6.}  % ????????
\caption{$F$ comparison, $\varepsilon=0.009$, $\sigma=10$, $N=128$.}
   \begin{tabular}{|c|c|c|c|c|c|}  % ?????
     \hline
       $F$ & 64 & 128 & 256 & 512 & 1024 \\   % ?????п?
     \hline
        Acc \%& 93.44 & 93.03 & \textbf{93.80} & 93.07 & 93.19\\
%        Acc \%& 93.4360 & 93.0308 & \textbf{93.8006} & 93.0713 & 93.1929\\
     \hline
   \end{tabular}
\label{table:t2}
\end{table}

%The compared methods include 3DGAN~\cite{3dganWu}, PointNet++~\cite{nipspoint17}, FoldingNet~\cite{YaoqingCVPR2018}, PANORAMA~\cite{Sfikas17}, Pairwise~\cite{JohnsLD16}, GIFT~\cite{tmmbs2016}, DominantSet~\cite{chuwang2017}, MVCNN~\cite{su15mvcnn}, SphericalProjection~\cite{huang2017spherical}, RotationNet~\cite{AsakoCVPR2018}, SO-Net~\cite{Jiaxincvpr18}, SVSL~\cite{Zhizhong2018seq}, and VIPGAN~\cite{Zhizhong2018VIP}.

\noindent\textbf{Parameters. }We first explore how the important parameters $F$, $N$ and $\sigma$ affect the performance of 3DViewGraph under ModelNet40. The comparison in Table.~\ref{table:t2},~\ref{table:t3}, and ~\ref{table:t3sigma} shows that their effects are slight in a proper range.

\begin{table}[h]
\centering
%\caption{The number of row-wise kernels and droupout ratio comparison under ModelNet40, $\varepsilon$=2e-6.}  % ????????
\caption{$N$ comparison, $\varepsilon=0.009$, $\sigma=10$, $F=256$.}
   \begin{tabular}{|c|c|c|c|c|c|}  % ?????
     \hline
       $N$ & 32 & 64 & 128 & 256 & 512 \\   % ?????п?
     \hline
        Acc \%& 90.84 & 92.91 & \textbf{93.80} & 93.44 & 93.40\\
%        Acc \%& 90.8428 & 92.9092 & \textbf{93.8006} & 93.4360 & 93.3955\\
     \hline
   \end{tabular}
\label{table:t3}
\end{table}

\noindent\textbf{Classification. }As compared under ModelNet in Table~\ref{table:t5}, 3DViewGraph outperforms all the other methods under the same condition\footnote{\noindent We use the same modality of views from the same camera system for the comparison, where the results of RotationNet are from Fig.4 (d) and (e) in https://arxiv.org/pdf/1603.06208.pdf. Moreover, the benchmarks are with the standard training and test split.}. In addition, we show the single view classification accuracy in VGG fine-tuning (``VGG(ModelNet)''). To highlight the contribution of VGG fine-tuning, spatial similarity, and attention, we remove fine-tuning (``Ours(No finetune)'') or set all spatial similarity (``Ours(No spatiality)'') and attention (``Ours(No attention)'') to 1. The degenerated results indicate these elements are important for 3DViewGraph to achieve high accuracy. Similar phenomena is observed when we justify the effect of $\bm{C}^i_j$ and $\bm{W}_F$ in Eq.~\ref{eq:attention} by setting them to 1 (``Ours(No attention-)''), respectively. We also justify the latent semantic embedding and spatial pattern correlation by replacing them by single view features (``Ours(No latent)'') and summation (``Ours(No correlation)''), the degenerated results also show that they are important elements. Finally, we compare our proposed view aggregation with mean (``Ours(MeanPool)'') and max pooling (``Ours(MaxPool)'') by directly pooling all single view features together. Due to the loss of content information in each view and spatial information among multiple views, pooling performs worse.

\begin{table}[h]
\centering
%\caption{The number of row-wise kernels and droupout ratio comparison under ModelNet40, $\varepsilon$=2e-6.}  % ????????
\caption{$\sigma$ comparison, $\varepsilon=0.009$, $N=128$, $F=256$.}
   \begin{tabular}{|c|c|c|c|c|c|}  % ?????
     \hline
       $\sigma$ & 0 & 1 & 5 & 10 & 11\\   % ?????п?
     \hline
        Acc \%& 92.91 & 93.48 & 93.72 & \textbf{93.80} & 93.48 \\
     \hline
   \end{tabular}
\label{table:t3sigma}
\end{table}

\begin{table}[h]
\centering
\caption{Classification comparison under ModelNet with $\varepsilon=0.009$, $\sigma=10$, $F=256$, $N=128$, unless noted otherwise.}  % ????????
    \begin{tabular}{|c|c|c|}  % ?????
     \hline
          Methods & MN40(\%) & MN10(\%) \\   % ?????п?
     \hline
       %SHD\cite{Kazhdan03} & Mesh & - & 68.23 & - \\
%       LFD\cite{Chen03} & Image & 10 & 75.47 & - \\
%       PyramidHoG-LFD & Image & 20 & 87.2 & 90.5 \\
%       Fisher vector\cite{su15mvcnn}& - & 12 & 84.8 & - \\
%       3DShapeNets\cite{Wu2015} & Voxel & 12 & 77.32 & - \\
%       DeepPano\cite{Bshi2015} & Image & 1 & 77.6 & - \\
%       Geometry image\cite{eccvSinha2017} & Image & 1 & 83.9 & - \\
%       VoxNet\cite{Maturana15} & Voxel & - & 83.0 & - \\
%       VRN\cite{Brocknips2016}& Voxel & 24 & - & 91.33 \\
%       FPNN\cite{LiPSQG16}& Voxel & - & 88.4 & - \\
%       T-L Network\cite{Girdhar16} & Voxel & - & 74.4 & - \\
       3DGAN\cite{3dganWuijcai} & 83.3 & 91.0 \\
%       PointNet\cite{cvprpoint2017} & Point & 1 & 86.2 & 89.2 \\
	   PointNet++\cite{nipspoint17ijcai} & 91.9 & - \\
	   FoldingNet\cite{YaoqingCVPR2018} & 88.4 & 94.4 \\
       %Octree  & 90.6 & - \\
       PANO\cite{Sfikas17ijcai} & 90.7 & 91.1 \\
       Pairwise\cite{JohnsLD16} & 90.7 & 92.8 \\
       GIFT\cite{tmmbs2016ijcai}  & 89.5 & 91.5 \\
       Domi\cite{chuwang2017ijcai}  & 92.2 & - \\
       MVCNN\cite{su15mvcnnijcai}  & 90.1 & - \\
       %MVCNN\cite{su16mvcnn} & Image & 20 & 89.7 & 92.0\\
%       MVCNN-Sphere\cite{su16mvcnn} & Voxel & 20 & 86.6 & 89.5 \\
       Spherical\cite{huang2017spherical} & 93.31 & -\\
       Rotation\cite{AsakoCVPR2018}  & 92.37 & 94.39 \\
       SO-Net\cite{Jiaxincvpr18ijcai}  & 90.9 & 94.1 \\
       %MHBN  & 92.2 & 94.1 \\
       SVSL\cite{Zhizhong2018seqijcai} & 93.31 & 94.82 \\
       VIPGAN\cite{Zhizhong2018VIP}& 91.98 & 94.05 \\
     \hline
       VGG(ModelNet40)  & 87.27 & - \\
       VGG(ModelNet10)  & - & 88.63\\
       %VGG(Voting) & Image & 20 & 90.54 & 93.15 \\
       Ours  & \textbf{93.80} & \textbf{94.82}\\
       %Ours1 & \textbf{91.47} & \textbf{93.35} \\
       Ours($\sigma=5$)  & \textbf{93.72} & \textbf{95.04} \\
       %Ours1($\sigma=5$)  & \textbf{91.49} & \textbf{93.07} \\
       Ours(No finetune) &  90.40 & - \\
       Ours(No spatiality)  & 92.91 & 94.16 \\
       Ours(No attention)  & 93.07 & 93.72 \\
       %Ours(No attention,$\sigma=5$) & 93.11 & 93.83 \\
       Ours(No attention-$\bm{C}^i_j$) & 91.82 & 93.39 \\
       Ours(No attention-$\bm{W}_F$)& 91.57 & 93.28 \\
       Ours(No latent) & 92.34 & 92.95 \\
       Ours(No correlation) & 89.30 & 93.83 \\
       %Ours(No finetune,$\sigma=5$)  & 87.16 & 90.15 \\
       Ours(MeanPool)  & 92.38 & 93.06\\
       Ours(MaxPool)& 91.89 & 92.84 \\
     \hline
   \end{tabular}
   \label{table:t5}
\end{table}

3DViewGraph also achieves the best under the more challenging benchmark ShapeNetCore55, based on the fine-tuned VGG (``VGG(ShapeNetCore55)''), as shown in Table~\ref{table:t7}. We also find that different parameters do not significantly affect the performance, such as $N$ and $\sigma$.

\begin{table}[h]
\centering
%\caption{Classification comparison under ShapeNetCore55, $K$=512, $\varepsilon$=4e-6.}  % ????????
\caption{Classification comparison under ShapeNetCore55 with $\varepsilon=0.009$, $\sigma=10$, $F=256$, $N=128$, unless noted otherwise.}  % ????????
    \begin{tabular}{|c|c|c|}  % ?????
     \hline
          Methods & Views & Accuracy(\%) \\
     \hline
       VIPGAN\cite{Zhizhong2018VIP} & 12 & 82.97 \\
       SVSL\cite{Zhizhong2018seqijcai} & 12 & 85.47 \\
     \hline
       VGG(ShapeNetCore55)  & 1 & 81.33\\
       %VGG(Voting) & Image & 20 & 74.45 & 86.19 \\%????voting
       %VGG(Voting) & Image & 20 & 74.63 & 86.40 \\%????voting
       Ours  & 20 &   \textbf{86.87}\\
       %Ours1 & Image & 20 & \textbf{75.60} & 86.33\\
       Ours($N=256$) & 20  &  \textbf{86.36}\\
       %Ours1($N=256$) & Image & 20 & \textbf{74.57} &  86.16\\
       Ours($\sigma=5$)  & 20  &  \textbf{86.56}\\
       %Ours1($\sigma=5$) & Image & 20 & \textbf{75.56} &  86.36\\
       Ours($\sigma=5$,$N=256$)  & 20  &  \textbf{86.71}\\
       %Ours1($\sigma=5$,$N=256$) & Image & 20 & \textbf{75.42} &  86.10\\
     \hline
   \end{tabular}
   \label{table:t7}
\end{table}

\begin{figure}[h]
  \centering
  % the following command controls the width of the embedded PS file
  % (relative to the width of the current column)
  %\includegraphics[width=.95\linewidth, bb=39 696 126 756]{figures/definition3.eps}
   \includegraphics[width=\linewidth]{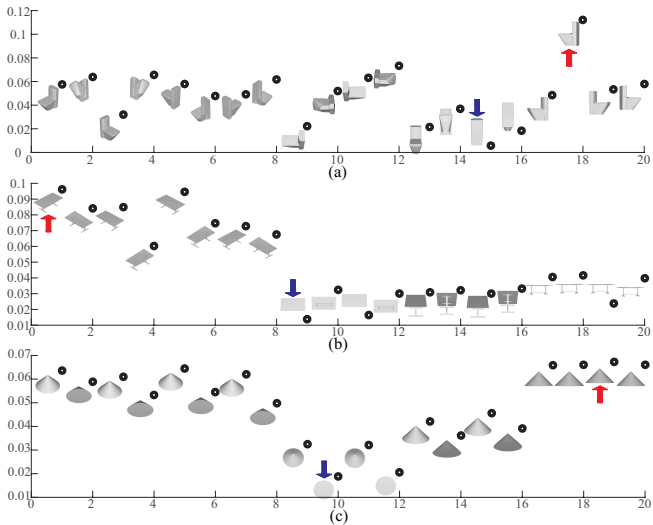}
  % replacing the above command with the one below will explicitly set
  % the bounding box of the PS figure to the rectangle (xl,yl),(xh,yh).
  % It will also prevent LaTeX from reading the PS file to determine
  % the bounding box (i.e., it will speed up the compilation process)
  % \includegraphics[width=.95\linewidth, bb=39 696 126 756]{sampleFig}
  %
  %
\caption{\label{fig:att}The visualization of attention weights (black nodes) learned for views of (a) a toilet, (b) a table, and (c) a cone. The highest and lowest attention weights are indicated by red upward arrow and blue downward arrow, respectively.}
\end{figure}

\noindent \textbf{Attention visualization. }We visualize the attention learned by 3DViewGraph under ModelNet40, which demonstrates how 3DViewGraph understands 3D shapes by analyzing views on a view graph. In Fig.~\ref{fig:att}, attention weights $\bm{\alpha}^i$ on view nodes $D_j^i$ of $G^i$ are visualized as a vector which is represented by scattered black nodes, where the corresponding views are also shown nearby, such as the views of a toilet in Fig.~\ref{fig:att}(a), a table in Fig.~\ref{fig:att}(b) and a cone in Fig.~\ref{fig:att}(c). The coordinates of black nodes along the y-axis indicate how much attention 3DViewGraph pays to the corresponding view nodes. In addition, the views that is paid the most and least attention to are highlighted by the red upward and blue downward arrow, respectively.

\begin{table*}[!htb]
\centering
\caption{Retrieval comparison under ShapeNetCore55, $\varepsilon=0.009$, $\sigma=10$, $F=256$, $N=128$.}  % ????????
    \begin{tabular}{|c|c|c|c|c|c|c|c|c|c|c|c|}  % ?????
     \hline
       \multirow{2}{*}{} & \multicolumn{5}{|c|}{micro} & \multicolumn{5}{|c|}{macro} \\
     \hline
        Methods & P@N & R@N & F1@N & mAP@N & NDCG@N & P@N & R@N & F1@N & mAP@N & NDCG@N \\  % ?????п?
     \hline
       %\multirow{10}{*}{Tesing} & \multirow{10}{*}{} & \multicolumn{5}{|c|}{\multirow{10}{*}{}} & \multicolumn{5}{|c|}{\multirow{10}{*}{}}\\
       %\multirow{10}{*}{Tesing} & & & & & & & & & & & \\
       % Kanezaki & 0.810 & 0.801 & \textbf{0.798} & 0.772 & 0.865 & 0.602 & 0.639 & \textbf{0.590} & 0.583& 0.656\\
%       Zhou & 0.786 & 0.773 & 0.767 & 0.722 & 0.827 & 0.592 & 0.654 & 0.581 & 0.575& 0.657 \\
%       Tatsuma  & 0.765 & 0.803 & 0.772 & 0.749 & 0.828 & 0.518 & 0.601 & 0.519 & 0.496 & 0.559 \\
%       Furuya  & \textbf{0.818} & 0.689 & 0.712 & 0.663 & 0.762 & \textbf{0.618} & 0.533 & 0.505 & 0.477 & 0.563 \\
%       Thermos  & 0.743 & 0.677 & 0.692 & 0.622 & 0.732 & 0.523 & 0.494 & 0.484 & 0.418 & 0.502 \\
%       Deng  & 0.418 & 0.717 & 0.479 & 0.540 & 0.654 & 0.122 & 0.667 & 0.166 & 0.339 & 0.404 \\
%       Li  & 0.535 & 0.256 & 0.282 & 0.199 & 0.330 & 0.219 & 0.409 & 0.197 & 0.255 & 0.377 \\
%       Mk  & 0.793 & 0.211 & 0.253 & 0.192 & 0.277 & 0.598 & 0.283 & 0.258 & 0.232 & 0.337 \\
%       Su  & 0.770 & 0.770 & 0.764 & 0.735 & 0.815 & 0.571 & 0.625 & 0.575 & 0.566 & 0.640 \\
%       Bai  & 0.706 & 0.695 & 0.689 & 0.640 & 0.765 & 0.444 & 0.531 & 0.454 & 0.447 & 0.548 \\
       All & \textbf{0.818} & 0.803 & \textbf{0.798} & 0.772 & 0.865 & \textbf{0.618} & 0.667 & \textbf{0.590} & 0.583 & 0.657 \\
	   Taco  & 0.701 & 0.711 & 0.699 & 0.676 & 0.756 & - & - & - & - & - \\
     \hline
       Ours  & 0.6090 & \textbf{0.8034} & 0.6164 & \textbf{0.8492} & \textbf{0.9054} & 0.1929 & \textbf{0.8301} & 0.2446 & \textbf{0.7019} & \textbf{0.8461} \\%@715
     \hline
     %\multirow{6}{*}{Validation}& Su & 0.805 & 0.800 & \textbf{0.798} & 0.910 & 0.938 & 0.641 & 0.671 & \textbf{0.642} & 0.879& 0.920\\
%       & Bai & 0.747 & 0.743 & 0.736 & 0.872 & 0.929 & 0.504 & 0.571 & 0.516 & 0.817& 0.889 \\
%       & Li  & 0.343 & \textbf{0.924} & 0.443 & 0.861 & 0.930 & 0.087 & \textbf{0.873} & 0.132 & 0.742 & 0.854 \\
%       & Wang  & 0.682 & 0.527 & 0.488 & 0.812 & 0.881 & 0.247 & 0.643 & 0.266 & 0.575 & 0.712 \\
%       & Tatsuma  & 0.306 & 0.763 & 0.378 & 0.722 & 0.886 & 0.096 & 0.828 & 0.140 & 0.601 & 0.801 \\
%       & Ours  & \textbf{0.8737} & 0.1233 & 0.1731 & \textbf{0.9482} & \textbf{0.9526} & \textbf{0.6428} & 0.3935 & 0.3833 & \textbf{0.9054} & \textbf{0.9334} \\%@18
%       \hline
%     \multirow{6}{*}{Training}& Su & 0.939 & 0.944 & \textbf{0.941} & 0.964 & 0.923 & 0.909 & 0.935 & \textbf{0.921} & 0.964 & 0.947\\
%       & Bai & 0.841 & 0.571 & 0.620 & 0.907 & 0.912 & 0.634 & 0.452 & 0.472 & 0.815 & 0.891 \\
%       & Li  & 0.827 & \textbf{0.996} & 0.864 & 0.990 & 0.978 & 0.374 & \textbf{0.997} & 0.460 & 0.982 & 0.986 \\
%       & Wang  & 0.884 & 0.260 & 0.363 & 0.917 & 0.891 & 0.586 & 0.497 & 0.428 & 0.775 & 0.863 \\
%       & Ours  & \textbf{0.9952} & 0.0058 & 0.0115 & \textbf{0.9995} & \textbf{0.9847} & \textbf{0.9931} & 0.0220 & 0.0424 & \textbf{0.9994} & \textbf{0.9909} \\%@3
%     \hline
   \end{tabular}
   \label{table:t10}
\end{table*}

\begin{table}[h]
\centering
%\caption{Retrieval comparison (mAP) under ModelNet40 and ModelNet10, $K$=512, $\varepsilon$=4e-6.}  % ????????
\caption{Retrieval comparison (mAP) under ModelNet, $\varepsilon=0.009$, $\sigma=10$, $F=256$, $N=128$.}  % ????????
\resizebox{0.5\textwidth}{!}{
    \begin{tabular}{|c|c|c|c|}  % ?????
     \hline
       Methods & Range & MN40 & MN10 \\  % ?????п?
     \hline
       SHD & Test-Test & 33.26 & 44.05 \\
       LFD & Test-Test & 40.91 & 49.82 \\
       3DNets\cite{Wu2015ijcai} & Test-Test & 49.23 & 68.26 \\
       GImage\cite{eccvSinha2017} & Test-Test & 51.30 & 74.90 \\
       DPano\cite{Bshi2015ijcai}& Test-Test & 76.81 & 84.18 \\
       MVCNN\cite{su15mvcnnijcai} & Test-Test & 79.50 & - \\
       PANO\cite{Sfikas17ijcai} & Test-Test & 83.45 & 87.39 \\
       GIFT\cite{tmmbs2016ijcai} & Random & 81.94 & 91.12\\
       Triplet\cite{Xinweicvpr18}& Test-Test & 88.0 & - \\
     \hline
       Ours & Test-Test & \textbf{90.54} & \textbf{92.40}\\
       Ours & Test-Train & \textbf{93.49} & \textbf{95.17}\\
       Ours & Train-Train & \textbf{98.75} & \textbf{99.79}\\
       Ours & All-All & \textbf{96.95} & \textbf{98.52}\\
     \hline
   \end{tabular}}
   \label{table:t9}
\end{table}

\begin{wrapfigure}{r}{0.53\linewidth}% 靠文字内容的左侧
\includegraphics[width=\linewidth]{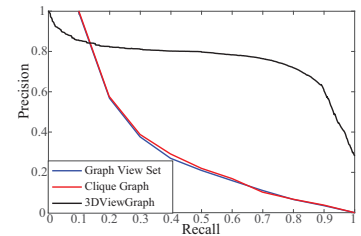}
\caption{The precision and recall comparison with graph-based multi-view learning methods under PSB.}\label{fig:PSB}
\end{wrapfigure}

Fig.~\ref{fig:att} demonstrates that 3DViewGraph is able to understand each view, since the view with the most ambiguous appearance in a view graph is depressed while the view with the most distinctive appearance is highlighted. For example, the most ambiguous views of toilet, table and cone merely show some basic shapes %without any clue which are impossible to be classified,
that provide little useful information for classification, such as the rectangles of the toilet and table, and the circle of the cone. In contrast, the most distinctive views of toilet, table and cone exhibit more unique and distinctive characteristics.

\begin{figure}[h]
  \centering
  % the following command controls the width of the embedded PS file
  % (relative to the width of the current column)
  %\includegraphics[width=.95\linewidth, bb=39 696 126 756]{figures/definition3.eps}
   \includegraphics[width=\linewidth]{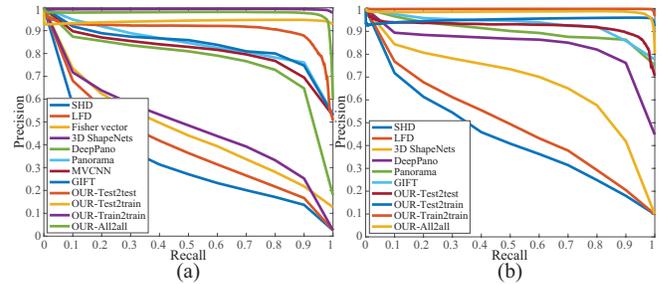}
  % replacing the above command with the one below will explicitly set
  % the bounding box of the PS figure to the rectangle (xl,yl),(xh,yh).
  % It will also prevent LaTeX from reading the PS file to determine
  % the bounding box (i.e., it will speed up the compilation process)
  % \includegraphics[width=.95\linewidth, bb=39 696 126 756]{sampleFig}
  %
  %
\caption{\label{fig:PR} The precision and recall cure comparison among different methods under (a) ModelNet40 and (b) ModelNet10.}
\end{figure}

\noindent\textbf{Retrieval. }We evaluate the retrieval performance of 3DViewGraph under ModelNet in Table~\ref{table:t9}. We outperform the state-of-the-art methods, where the retrieval range is also shown. We further detail the precision and recall curves of these results in Fig.~\ref{fig:PR}. In addition, 3DViewGraph also achieve the best results under ShapeNetCore55 in Table~\ref{table:t10}. We compare 10 state-of-the-art methods under testing set in the SHREC2017 retrieval contest~\cite{3dor20171050ijcai} and Taco~\cite{s2018spherical}, where we summarize all the 10 methods (``All'') by presenting the best result of each metric due to page limit. Finally, we demonstrate that 3DViewGraph is also superior to other graph-based multi-view learning methods~\cite{Liu2015GCV,AnnLiu16} under Princeton Shape Benchmark (PSB) in Fig.~\ref{fig:PSB}.

%including 3DShapeNets~\cite{Wu2015}, Geometry image~\cite{eccvSinha2017}, DeepPano~\cite{Bshi2015}, MVCNN~\cite{su15mvcnn}, PANORAMA~\cite{Sfikas17}, GIFT~\cite{tmmbs2016}, Triplet-Center~\cite{Xinweicvpr18}

\section{Conclusion}
\label{sec:conclusion}
In view-based deep learning models for 3D shape analysis,  view aggregation via widely used pooling, leads to information loss about content and spatial relationship of views. We propose 3DViewGraph to address this issue for 3D global feature learning by more effectively aggregating unordered views with attention. By organizing unordered views taken around a 3D shape into a view graph, 3DViewGraph learns global features of the 3D shape
%from the view graph
by simultaneously encoding both the content information within view nodes and the spatial relationship among the view nodes. Through a novel latent semantic mapping, low-level view features are projected into a meaningful, lower-dimensional latent semantic embedding using a learned kernel function, which directly captures the similarities between low-level view features and latent semantic patterns. The latent semantic mapping successfully facilitates 3DViewGraph to encode the content information and the spatial relationship in each pair of view nodes using a novel spatial pattern correlation. Further, our novel attention mechanism effectively increases the discriminability of learned features by efficiently highlighting the unordered view nodes with distinctive characteristics and depressing the ones with appearance ambiguity. Our results in classification and retrieval under three large-scale benchmarks show that 3DViewGraph can learn better global features than the state-of-the-art methods due to its more effective view aggregation.

\section{Acknowledgments}
This work was supported by National Key R\&D Program of China (2018YFB0505400), NSF (1813583), University of Macau (MYRG2018-00138-FST), and FDCT (273/2017/A). We thank all anonymous reviewers for their constructive comments.

\bibliographystyle{named}
\bibliography{paper}

\end{document}